\pdfoutput=1

\documentclass[11pt]{article}

\usepackage[]{acl}

\usepackage{times}
\usepackage{latexsym}
\usepackage{framed}
\usepackage{amsmath} 
\usepackage{amssymb}
\usepackage{graphicx}
\usepackage{url}
\usepackage{tabularx}
\usepackage{booktabs}
\usepackage{multirow}
\usepackage{soul}
\usepackage{multirow}

\usepackage[T1]{fontenc}

\usepackage[utf8]{inputenc}
\usepackage{microtype}

\newcommand*{\affaddr}[1]{#1}

\title{Quiz Design Task:\\Helping Teachers Create Quizzes with Automated Question Generation}

\author{
  \quad \textbf{Philippe Laban}
  \quad \textbf{Chien-Sheng Wu}
  \quad \textbf{Lidiya Murakhovs'ka} \\
   \quad \textbf{Wenhao Liu}
  \quad \textbf{Caiming Xiong} \\
  \affaddr{Salesforce AI Research} \\
  \{plaban, wu.jason, l.murakhovska, wenhao.liu, cxiong\}@salesforce.com
}

\begin{document}
\maketitle
\begin{abstract}
Question generation (QGen) models are often evaluated with standardized NLG metrics that are based on n-gram overlap.
In this paper, we measure whether these metric improvements translate to gains in a practical setting, focusing on the use case of helping teachers automate the generation of reading comprehension quizzes. In our study, teachers building a quiz receive question suggestions, which they can either accept or refuse with a reason. Even though we find that recent progress in QGen leads to a significant increase in question acceptance rates, there is still large room for improvement, with the best model having only 68.4\% of its questions accepted by the ten teachers who participated in our study. We then leverage the annotations we collected to analyze standard NLG metrics and find that model performance has reached projected upper-bounds, suggesting new automatic metrics are needed to guide QGen research forward.
\end{abstract}

\section{Introduction}

Question generation is a text generation task with practical applications in several settings such as asking clarification questions in dialogue systems \cite{braslavski2017you}, recommending questions during a reading session \cite{laban2020s}, or other educational scenarios such as creating quizzes to emphasize core concepts and engage learners through interaction \cite{kurdi2020systematic, steuer2021not}.

\begin{figure}
    \centering
    \includegraphics[width=0.42\textwidth]{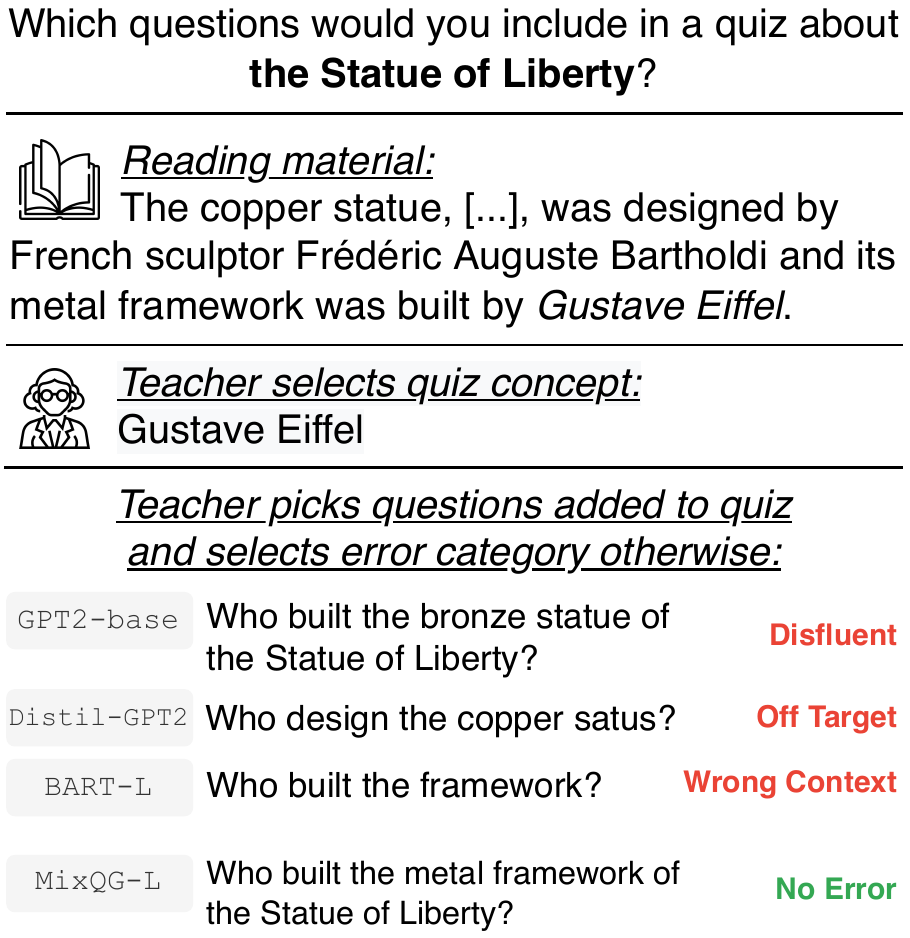}
    \caption{\textbf{Illustration of the Quiz Design Task.} For a topic, a teacher selects a quiz concept, picks which candidate questions from various models to include in the quiz, and gives a reason to reject others.}
    \label{fig:intro_qge}
\end{figure}

The most common automatic evaluation of QGen borrows from other NLG tasks, using metrics such as BLEU \cite{papineni2002bleu} to compare system-generated questions with held-out human-written references in terms of n-gram overlap \cite{amidei2018evaluation}. Although they are straightforward to compute, these metrics have been shown to correlate weakly with human opinion in NLG \cite{gatt2018survey}, do not provide a ceiling performance, or insights into the types of errors prevalent in generated questions.

Some prior work has proposed automatic metrics that are specific to QGen, however the metrics are either rule-based \cite{nema2018towards}, matching for the presence of certain elements in generated question with limited flexibility, or shown not be beneficial when used to optimize a QGen model through Reinforcement Learning, according to human raters \cite{hosking2019evaluating}.

In this paper, we propose to evaluate QGen with the help of teachers through the \textbf{Quiz Design Task}, illustrated in Figure~\ref{fig:intro_qge}. Human teachers are tasked with creating reading comprehension quizzes for hypothetical students, and QGen models interactively suggest quiz questions which can be accepted or rejected by the teachers. Model performance is tied to the acceptance rate of each model, in other words, the best QGen model is the one with the largest proportion of accepted questions.

There are several definitions for QGen, from clarification question generation \cite{rao2018learning}, to knowledge-graph QGen \cite{indurthi2017generating}, multiple-choice distractor generation \cite{araki2016generating} and answer-aware QGen \cite{sun2018answer}, in which given a context paragraph and a target answer, the model must generate a question answered by the target. We select the answer-aware QGen setting for our evaluation, as it allows for teachers to guide the QGen model by selecting desired concepts to include in the quiz by selecting target answers.

Our contribution is threefold: 1) we propose the Quiz Design Task, a conceptually simple task that allows us to evaluate QGen models in the setting of helping teachers design quizzes. 2) We collect 3,164 human-annotated samples from running the Quiz Design Task with 10 teachers.
We find that acceptance rates of generated questions vary widely from as low as 30\% for small pre-trained Transformer models, up to 68\% for the best performing model we evaluated. 3) We carefully analyze annotator agreement levels and compare between our results and n-gram-based metrics, revealing that there is some correlation between the widely used metrics and model performance in the Quiz Design Task. We also report an estimate of a ceiling for these automatic scores, which are already neared by the state-of-the-art QGen models we evaluate. We release all annotations as well as the interface used during the study publicly.\footnote{\url{https://github.com/salesforce/QGen}}

\section{Quiz Design Task}

\begin{figure}
    \centering
    \includegraphics[width=0.4\textwidth]{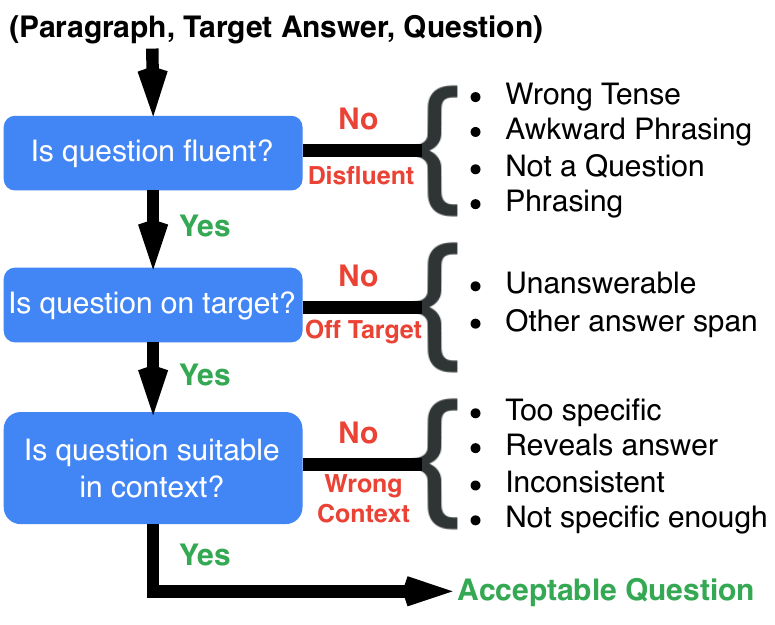}
    \caption{\textbf{Hierarchical categorization of errors for question generation.} Three error categories (Disfluent, Off Target, Wrong Context) each with several subtypes.}
    \label{fig:qgen_error_categorization}
\end{figure}

We propose to evaluate QGen models by measuring how helpful they are for quiz creation. Teachers often have experience with carefully crafting quiz questions, and possess knowledge as to what makes a quality question for a quiz \cite{pearson1983instruction, kendeou2016reading}. Meanwhile, they are for the most part unfamiliar with recent progress in language modeling, and do not necessarily know of the limitations of deep learning-based text generation. Therefore they can act as impartial judge in this particular setting in verifying whether question generation models have reached a level at which they can be used to facilitate reading comprehension quiz creation.

\subsection{Task Definition}

Teachers with experience in designing quizzes are invited to use a quiz design interface (Figure~\ref{fig:quiz_design_interface}), and follow the steps illustrated in Figure~\ref{fig:intro_qge}.
They begin by selecting a \textit{quiz topic}, such as the history of the Statue of Liberty in Figure~\ref{fig:intro_qge}. The system loads \textit{reading material} relevant to the topic, which can be sourced from a textbook or Wikipedia.

The objective for the teacher is to leverage the reading material and automated QGen models to design an entire quiz composed of 8-12 questions. The teachers proceed by selecting a \textit{quiz concept}, such as an entity, phrase, or keyword they wish to probe students on. Each evaluated QGen model then generates a candidate question given the entire reading material and the selected quiz concept.

After receiving candidate questions from the QGen models, teachers review and pick which to include in the quiz. Importantly, candidate questions are anonymized and presented in a shuffled order.
It is possible that several QGen models generate identical candidates, so we deduplicate the candidates before presenting them to annotators.

Existing question answering human evaluation design either automatically select quiz concepts or answers and questions are evaluated by distinct crowd-workers \cite{du2017learning, trischler2017newsqa}. In the case of Quiz Design Task, we believe that it is important to enable teachers to select quiz concepts themselves, as it allows them to have specific learning objectives, permitting them to assess generated questions with this context in mind.

\subsection{Question Error Categorization}

To understand model performance beyond overall acceptance rates and assess model limitations, annotators were made to select a reason for each rejected question. However, unlike other NLG tasks, QGen does not have an established error categorization.
Therefore, we carried out a formative study to construct a reusable error categorization for QGen. We collected questions by sampling the QGen models used in the study, and gradually constructed the categorization by labeling and refining the annotations on 976 generated questions. The final categorization is illustrated in Figure~\ref{fig:qgen_error_categorization}.

The QGen error categorization we propose is hierarchical, with errors falling in three nested categories. First, similar to the MQM categorization \cite{lommel2014multidimensional} used for translation, the question can be rejected because it is \textit{disfluent} for example with errors in grammar or repetition. Second, if the question is fluent, it can be rejected for being \textit{off target}: the answer to the generated question is not the target answer originally selected. Third, if the question is fluent and on target, it can be rejected for being wrong in context (\textit{wrong context}), for example by being too specific to be natural or not specific enough to be self-contained. Examples of question errors in each category in Table~\ref{table:categorization_examples}. 

\section{Quiz Setup and Results}

\subsection{Participant Recruitment}

We recruit teachers or ex-teachers from an online group forum. In total, 20 participants filled out an interest form, 14 were selected, and 10 completed the study (with the other 4 either forgetting to complete the task, or completing it partially). The participants had been teachers for at least a year and 3.6 years on average, and had taught diverse subjects such as sciences, history, literature, and IT topics, at various levels from primary school to college-level. The study was meant to last a maximum of two hours, and participants were gifted a \$50 gift card upon completion.

The study session began with a tutorial on the interface (see Appendix~\ref{appendix:annotator_guidelines}) and detailed examples of the error categories. Participants could then clarify any detail before commencing annotation.

\subsection{Quiz Topic Selection}

Participants were tasked with creating between 5-7 quizzes, each with a minimum of 8 concepts, and could pick from a set list of 7 quiz topics, which we pre-selected from the list of featured Wikipedia articles\footnote{\url{https://en.wikipedia.org/wiki/Wikipedia:Featured_articles}}. We purposefully selected articles within different domains to benchmark the QGen models in diverse topical settings: two in physics (Sustainable Energy, Californium Atom), two in biology (DNA, Enzymes), two in history (Statue of Liberty, Palazzo Pitti), and one in geology (the K-T extinction). Participants were given the first 500 words of the Wikipedia page of each topic as reading material to select Quiz concepts from.

\subsection{QGen Models Evaluated}

\begin{figure}
    \centering
    \includegraphics[width=0.49\textwidth]{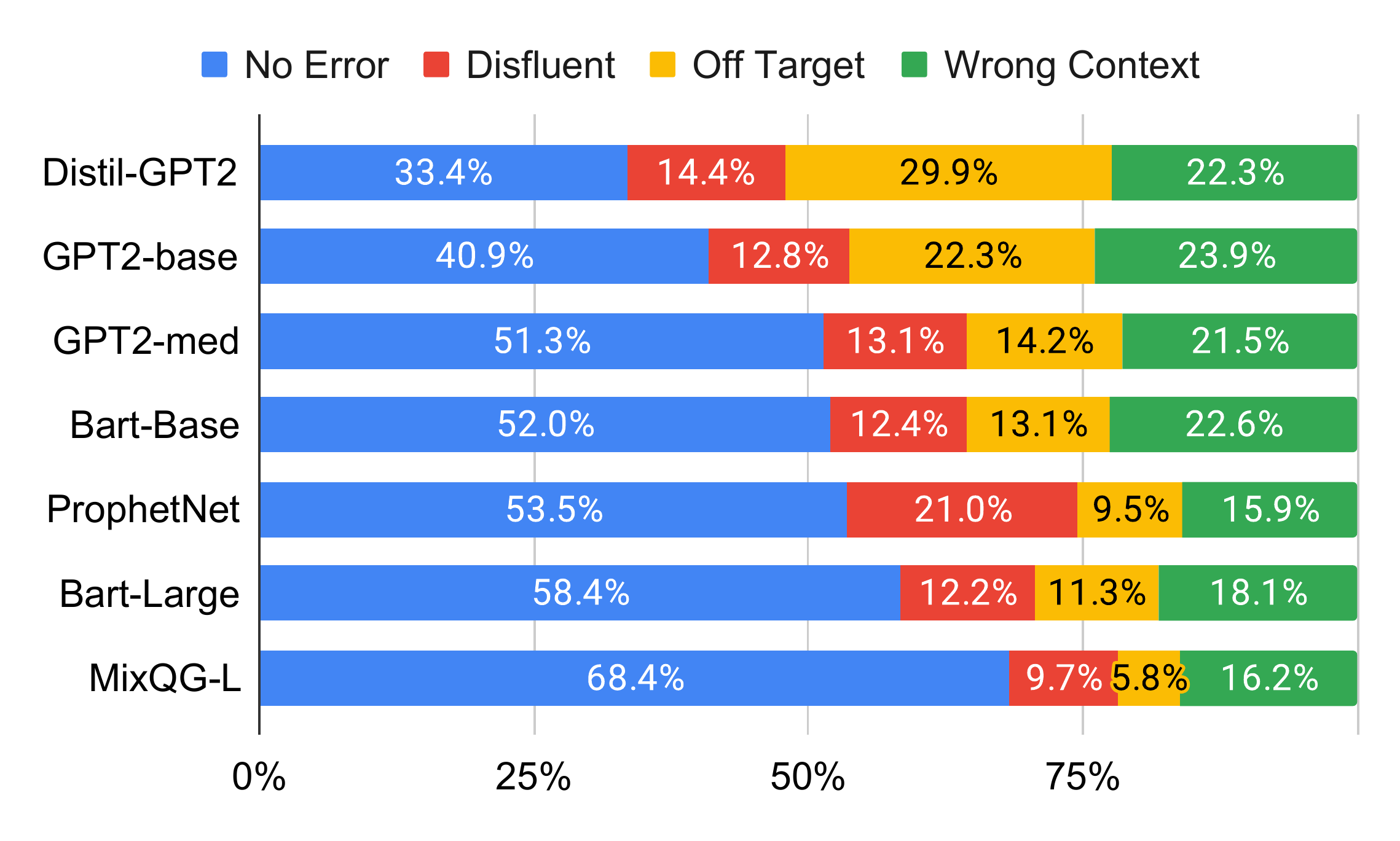}
    \caption{\textbf{Error distribution}. Seven QGen models are evaluated by 10 teachers on the Quiz Design Task. The high proportion of disfluency errors of ProphetNet is explained in Section~\ref{section:iaa}.}
    \label{fig:error_distribution}
\end{figure}

We include seven QGen models of varying size and architecture in our study. First, we finetune three GPT2 baselines \cite{radfordlanguage} on the SQuAD dataset \cite{rajpurkar2016squad}: \texttt{GPT2-distil} \cite{sanh2019distilbert}, \texttt{GPT2-base} and \texttt{GPT2-medium}. We further add two BART-based \cite{lewisbart} models trained on SQuAD as well: \texttt{BART-base} and \texttt{BART-large}. Finally, we include two recent QGen top-performers, \texttt{ProphetNet} \cite{qi2020prophetnet} and \texttt{MixQG-L} \cite{murakhovska2022mixqg}. We limit ourselves to seven models, and exclude larger models (such as GPT2-XL and MixQG-3b) to maintain an interface latency of under 200ms and limit burden to users \cite{miller1968response}. Details on model training and usage in Appendix~\ref{appendix:training_details}.

\subsection{Annotated Results}

In total, the study participants annotated 3,164 questions, with 52\% of them accepted into a quiz. The distribution of errors per model is summarized in Figure~\ref{fig:error_distribution}.
As expected, model size has an effect on performance, with the largest model MixQG-L achieving the highest performance with an acceptance rate of 68.4\%, which is more than double the 33.4\% achieved by Distil-GPT2.

Almost all models have the largest portion of errors coming from the Wrong Context category. In fact, model improvement mostly comes from the other two categories of errors, with a decrease of 40-80\% in numbers of errors made in the \texttt{Disfluent} and \texttt{Off Target} categories. In contrast, the MixQG model still generates a \texttt{Wrong Context} question 16.2\% of the time, a modest decrease from Distil-GPT2's 22.3\%.

As expected, the \texttt{Wrong Context} category is the most challenging: models have learned to generate fluent questions that are answered by a desired target concept, and still struggle with phrasing the question in a fashion adequate to the context.

\section{Analysis}

With the annotations collected, we calculate inter-annotator agreement and use the data to benchmark commonly-used NLG metrics.

\subsection{Inter-Annotator Agreement}
\label{section:iaa}

Even though we allow teachers to select their own quiz concepts, in 95 cases, two or more annotators selected the same concept and annotated an identical set of seven candidate questions. This leads us to have a total of 665 questions on which we can compute inter-annotator agreement. On this subset, we measure a Pearson correlation coefficient \cite{benesty2009pearson} of 0.47 which can be interpreted as moderate inter-annotator agreement \cite{schober2018correlation}.

When breaking down the analysis by model origin, the two lowest-performing models (Distil-GPT2 and GPT2-base) obtain the highest agreement rates (above 0.6), showing a stronger agreement on low-quality questions.
Notably, ProphetNet obtained the lowest agreement level (0.26).
Further investigation reveals that it is the only model generating questions in lowercase. Because our guidelines did not specify how to deal with improper capitalization, some annotators labeled lower-cased questions as a fluency error. This further explains why ProphetNet generated the largest number of disfluent questions. Future work should carefully indicate how to deal with casing and other normalization (such as punctuation) errors.

\subsection{Analysis of Existing Metrics}

\begin{table}[]
    \resizebox{0.50\textwidth}{!}{%

    \begin{tabular}{lcccccc}
    \toprule{}
    \textbf{Model Name} & \textbf{\%Acc.} & \textbf{BLEU} & \textbf{R-1} & \textbf{R-L} & \textbf{MET} & \textbf{BERT} \\
    \cmidrule(r){1-2} \cmidrule(l){3-7}
    Distil-GPT2 & 33.4 & 21.2 & 47.4 & 45.4 & 36.8 & 50.2 \\
    GPT2-base & 40.9 & 26.3 & 53.1 & 51.1 & 43.0 & 56.1 \\
    GPT2-med & 51.3 & 31.2 & 57.6 & 55.4 & 46.1 & 59.5 \\
    BART-Base & 52.0 & 31.2 & 57.2 & 54.8 & 46.0 & 59.9 \\
    ProphetNet & 53.5 & 33.3 & 62.1 & 59.3 & 51.7 & 57.4 \\
    Bart-Large & 58.4 & 32.4 & 59.2 & 56.9 & 48.8 & 61.1 \\
    MixQG-L & 68.4 & 33.5 & 59.6 & 57.2 & 50.6 & 60.0 \\
    \cmidrule(r){1-2} \cmidrule(l){3-7}
    Upper Bound & 100.0 & 33.9 & 60.4 & 58.0 & 50.2 & 61.4 \\
    \cmidrule(r){1-2} \cmidrule(l){3-7}
    Instance Corr. & - & .201 & \textbf{.233} & .231 & .221 & \textbf{.244} \\
    System Corr. & - & \textbf{.724} & .665 & .672 & .689 & .711 \\
    \bottomrule{}
    \end{tabular}
    }
    \vspace{-5pt}
    \caption{\textbf{NLG evaluation metrics.} For each metric, an upper-bound, and correlations at the instance-level and system-level are computed.}
    \label{table:metrics_results}
\end{table}

Because several questions for each given context are annotated, we have a unique opportunity to study the commonly-used NLG metrics, and assess which correlate with our annotators' judgements.
We evaluate four of the most commonly used metrics in QGen evaluation: BLEU \cite{papineni2002bleu}, ROUGE \cite{lin2004rouge} (we include ROUGE-1 and ROUGE-L variants), METEOR \cite{banerjee2005meteor}, and BERTScore \cite{zhang2019bertscore}.  Results are detailed in Table~\ref{table:metrics_results}.

First, we can use accepted questions as references, and compute metric performance by each system on the dataset we've collected. For each metric, we can compute an instance-level correlation (i.e., how well does a metric correlate with annotations for each individual question), as well as system-level correlation (i.e. how similar is the ranking of models according to annotators and according to the metric). As echoed in previous work \cite{novikova2017we, chaganty2018price}, instance-level correlations are low, but the aggregated metric scores provide high correlation at the system level, with BLEU achieving the highest system-level correlation.

Second, in cases where several questions were marked as acceptable, we can consider each as a valid reference. In such a case, we generate all pairs of references, treating one as a candidate, the other as a reference and computing scores with  the standard metrics. The score obtained can be interpreted as an upper-bound for each metric, as they are scores obtained by questions that are judged to all be acceptable.

For all metrics, we find that MixQG has already either surpassed this upper-bound or is within 0.4-1.4 points of doing so.
This analysis reveals that even though standard metrics have been useful at measuring progress in NLG, upper-bound performance may be reached soon, and better metrics are needed to guide future progress in QGen and NLG research.

\section{Limitations}

We now discuss the limitations of the work we've presented.

First, even though we attempted to create a realistic scenario in which to evaluate QGen models, some components of the protocol are simplified for practical purposes. For example, the created quiz were not assigned to students, and we rely solely on the teacher's opinion of the questions as a signal of question quality. Pushing the study further by assigning the quizzes to students and tying question quality to student performance on the quiz would add complexity, but render the protocol more realistic and provide practical learning signals from students.

Second, although we treat teacher annotations as the ground truth, there is some level of disagreement amongst the teachers we recruited, and we measured a moderate level of agreement in Section~\ref{section:iaa}. This emphasizes the necessity of thorough and precise guidelines requirements for evaluation protocols, as our lack of rules around the treatment of capitalization of questions led to low agreement on questions generated by an uncased model.

Third, although we gathered a large number of annotations overall, with 3,164 questions annotated in total, this remains small due to the fact that there are many variables on which to break down performance on (e.g., source document, model of origin, annotator). We plan to release the annotation interface as well as the content and models we used to allow future work to expand and reproduce the results.

\section{Conclusion}

We introduce the Quiz Design task, a human evaluation protocol used to evaluate Question Generation models in an applied scenario. In the QD task, teachers creating a quiz for their students are recommended generated questions, which they can accept in their quiz or reject with a reason from a newly proposed error categorization.
We run a QD task with 10 teachers, annotating 3,164 questions originating from seven models, and find that acceptance rates vary widely with the latest QGen models obtaining the highest acceptance rate of 68.4\%.
Finally, analysis of automatic metrics on our task's data reveals that even though metrics correlate well with system-level ranks, models have reached potential metric upper-bounds, and improved metrics are required to guide NLG forward.

\section{Ethical Considerations}

Our experiments were all run for the English language, and even though we expect the study design to be adaptable to other languages, we have not verified this assumption experimentally and limit our claims to the English language. Expanding the claims to other languages would require trained question generation models in the studied language.

The teacher annotators that participated in our study were compensated at a rate above minimum wage, and we have insured that no personally identifiable information is available in the annotations we've released.

\bibliography{anthology,main}
\bibliographystyle{acl_natbib}

\clearpage
\appendix

\section*{Appendix}
\renewcommand{\thetable}{A\arabic{table}}
\setcounter{table}{0}
\renewcommand{\thefigure}{A\arabic{figure}}
\setcounter{figure}{0}

\section{Training Details}
\label{appendix:training_details}
We trained five of the QGen models used in the Quiz Design study. They were all trained for ten epochs on the training portion of the SQuAD dataset \cite{rajpurkar2016squad}, using the ADAM optimizer \cite{kingma2015adam}, with hyper-parameter tuning based on model loss on the validation set. The model checkpoint that achieves the lowest validation loss is selected as the final model. Selected hyper-parameters were:

\textbf{Distil-GPT2}: batch-size 32, learning rate $2*10^{-5}$.

\textbf{GPT2-base}: batch-size 32, learning rate $2*10^{-5}$.

\textbf{GPT2-medium}: batch-size 16, learning rate $2*10^{-5}$.

\textbf{BART-base}: batch-size 32, learning rate $1*10^{-4}$.

\textbf{BART-large}: batch-size 32, learning rate $2*10^{-5}$.

Finally, the last two QGen we used are publicly available on the HuggingFace model hub \cite{wolf2020transformers}, and we use them as is:

\textbf{ProphetNet}: \texttt{microsoft/ prophetnet-large- uncased-squad-qg}

\textbf{MixQG}: \texttt{Salesforce/mixqg-large}

With all models, we used beam search to generate candidate questions, using a beam-size of 2, and a sequence length maximum of 30.

\section{Guidelines to Annotators}
\label{appendix:annotator_guidelines}
We provide the exact guidelines that were given to study participants before they started the annotation procedure:

\begin{enumerate}
    \item Your objective is to design a quiz about a particular topic for a class of students. The procedure is the following:
    \item Select a quiz topic from the  list (for example "Sustainable Energy")
    \item The system will load a text about the topic.
    \item Select a concept that you want to quiz your students on (for example a phrase, a figure, or a keyword) and confirm your selection.
    \item \textbf{Important:} It is recommended to select \textbf{shorter concepts}, and not full sentences to obtain more precise question. Selecting concepts of up to about 8 words is ideal.
    \item The system will load a list of questions that attempt to quiz students about the selected concept.
    \item Go over each question, and remove ones you would not include in your quiz. We will next go over types of questions that should be removed.
    \item \textbf{Important:} you can keep one, multiple or none of the questions (if none of the questions are satisfactory). For each question you remove, you have to choose the reason that the question is unsatisfactory (more on this later).
    \item Once you've finalized the question for a concept, select another concept and repeat the question selection process. Try to select \textbf{8-12} concepts per topic to generate long enough quizzes.
    \item Once you've finished a full quiz set, you can move on to another quiz topic. We have found that in one hour, you should be able to complete the quizzes for 5 topics.
\end{enumerate}

Following these guidelines, the annotators were provided definitions for each error category, as well as examples similar to the ones shown in Table~\ref{table:categorization_examples}.

\section{Error Categorization Question Examples}
\label{appendix:categorization_examples}

The examples listed in Table~\ref{table:categorization_examples} were collected during a formative study to establish an error categorization for the task of Question Generation.

\begin{table*}[t]
    \resizebox{\textwidth}{!}{%
    \begin{tabular}{p{2.5cm} p{3cm} p{6cm} p{5cm}}
    \toprule{}
    \textbf{Category} & \textbf{Finer Category} & \textbf{Example Question} & \textbf{Rationale} \\
    \midrule{}
    \multirow{4}{*}[-1.3cm]{\Large{Disfluent}} & Wrong Tense & What were historically used to disenfranchise racial minorities? & Should be "What was historically..." \\
     & Awkward Phrasing & When did the woolly mammoth die? & Should be "go instinct" rather than "die" \\
     & Not a Question & In January 2020, scientists reported that climate-modeling of the extinction event favors the asteroid impact and not volcanism? & Sentence in declarative format \\
     & Repetition & Who led the team that led the K-Pg boundary clay? & "led" is repeated twice \\
    \midrule{}
    \multirow{2}{*}[-0.6cm]{\Large{Off Target}} & Unanswerable & Why are DNA studies so important? & Not answered in the DNA Wikipedia page. \\
     & Other Answer Span & Who designed the Statue of Liberty? & True answer is Bartholdi, even though target answer was Eiffel (the metalwork builder) \\
    \midrule{}
    \multirow{4}{*}[-1.3cm]{\Large{Wrong Ctxt}} & Too Specific & Where was the 181 km (114 mi) crater discovered? & Not standard to have unit translations in questions \\
     & Reveals Answer & What was the name of the Federal Reserve System? (leading to the creation of the Federal Reserve System) & Question's target answer is Federal Reserve System \\
     & Inconsistent & What are the only two animals that survived the Cretaceous-Paleogene extinction? & The Wikipedia article mentions species and not animals \\
     & Not Specific Enough & What are some ectothermic species? & Too many ectothermic species are mentioned in the article.\\
    \bottomrule{}
    \end{tabular}
    }
    \caption{\textbf{Example generated questions collected during formative study.} These examples form the basis for the error categorization we propose for the QGen task.}
    \label{table:categorization_examples}
\end{table*}

\section{Interface Screenshot}
\label{appendix:interface}

\begin{figure*}
    \centering
    \frame{\includegraphics[width=0.95\textwidth]{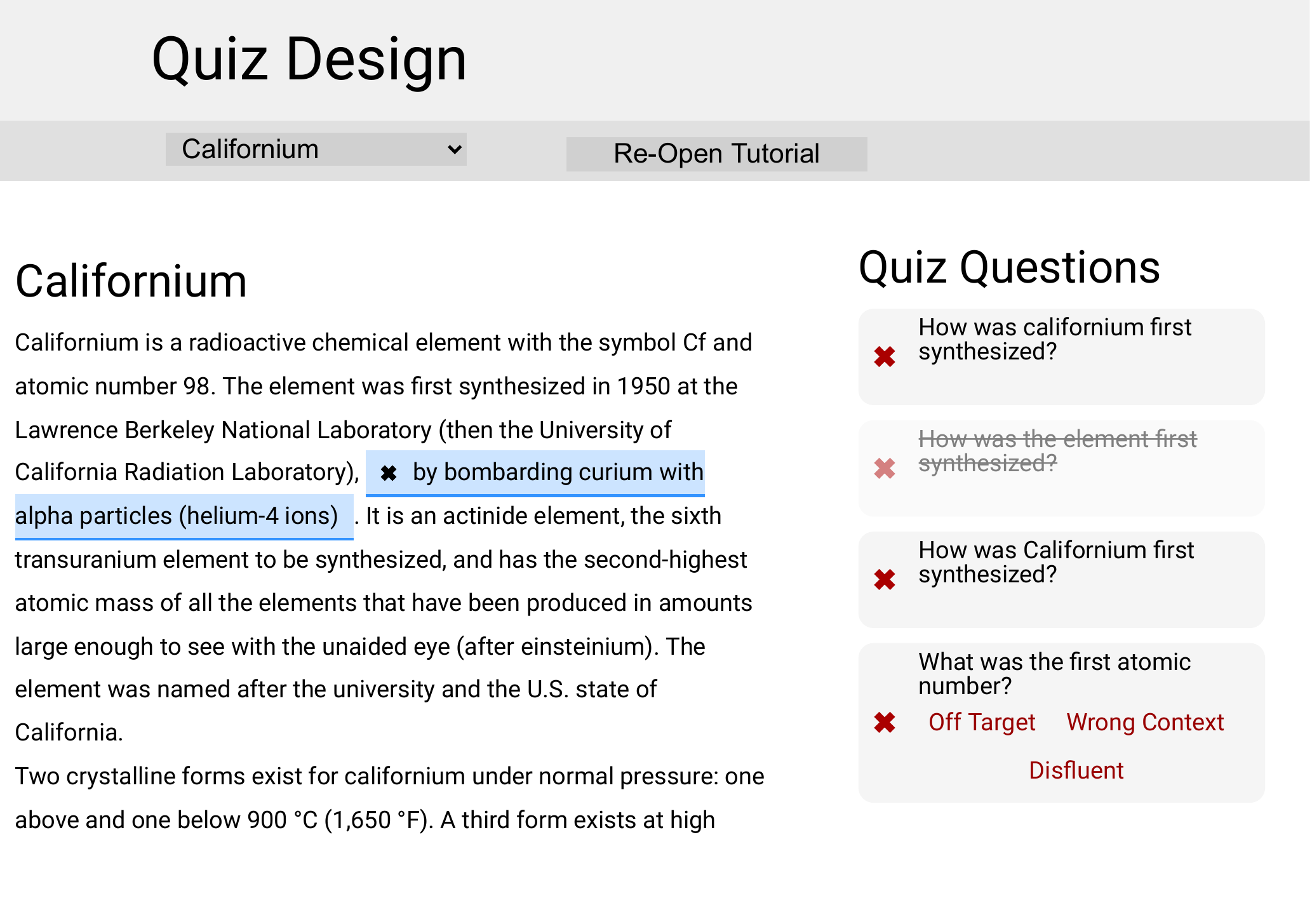}}
    \caption{\textbf{Screenshot of annotation interface used for the Quiz Design Task.} The teacher has selected the concept highlighted in blue in the reading material in the left column. In the right column, the system gives proposes candidate questions, which can be added to the quiz, or refused with a reason.}
    \label{fig:quiz_design_interface}
\end{figure*}

Figure~\ref{fig:quiz_design_interface} displays a screenshot of the interface used for the Quiz Design Task.

\end{document}